# A Multi-Tiered AI Accessibility Model (MAIAM): Proportionate Access and Capability Limits for Comprehensive Frontier AI Safety

Chai K. Toh and Vasu Raj Jain
SAFE AI Foundation USA
chai.toh@safeaifoundation.com, vasujain46@gmail.com

**Abstract**

Most debate about releasing capable AI systems treats the choice as binary: keep the model behind an API, or publish the weights. That framing runs together two separate decisions. One is how capable a system may become. The other is who may obtain a given capability, and on what terms. This paper is about the second decision. We propose a Multi-Tiered AI Accessibility Model (MAIAM) that sorts access by purpose, safety, and capability instead of by release artifact. Tier 1, AI for Learning, is broadly accessible and deliberately capability-bounded. Tier 2, AI for Exploration, serves research and development under registration, identity, and usage controls that tighten as capability rises. Tier 3, Full-Scale Frontier AI, is available only through structured, recorded, and accountable access. The principle behind the tiers is that accessibility should be proportional to capability, safety, and risk. We show why publishing frontier-class weights forfeits every safety lever a developer has after release, and, using evidence that safety fine-tuning can be stripped from open weights for a few hundred dollars, why an evaluation of the base model says little about its derivatives. We give criteria for assigning a system to a tier along capability, autonomy, domain-consequence, and usage/safety axes, and combine MAIAM with the SAFE AI Operating Point into a two-part framework: a Capability Boundary that fixes the safety limits not to be crossed, and an Accessibility Boundary that fixes who may obtain a capability below that limit. Capability limits alone do not make AI safe; a tiered access regime administered by regulating bodies is also needed.

## 1 Introduction

"AI" is no longer one kind of thing. The label covers a small instructional language model, a code assistant, a tool-using autonomous agent, and a frontier-scale reasoning model, and the worst that each of these can do differs by orders of magnitude. The release debate has not caught up with this. It still asks, in effect, whether AI should be open or closed, when the question that matters is whether every AI system, whatever its capability, should be equally accessible to everyone.

The usual way of answering is to argue about release mode. Developers of closed systems keep the weights and serve the model through a hosted interface. The open-weight ecosystem publishes weights that anyone can download, fine-tune, merge, quantize, wrap in an agent, and pass on. Both camps have real arguments, and neither framing separates the capability of a system from the question of who gets it.

We propose a Multi-Tiered AI Accessibility Model (MAIAM) in which access is a graduated variable, set by a system's capability, autonomy, safety, and domain of consequence. The model has three tiers, named for what they are for rather than for how the model is shipped: AI for Learning, AI for Exploration, and Full-Scale Frontier AI. Section 3 describes them. In what follows, "AI system" means both LLM models and the agents built on them.

### 1.1 Contributions and positioning

MAIAM is not a graduated release model. Release gradients [1], structured access [2], and staged-release proposals [3] all rejected the open/closed binary before us, but they describe choices a developer makes when shipping a model: which artifact to publish, and through which interface (weights, an API, keys). MAIAM sits above that. It is a user-side accessibility regime, a classification of who may download, access, and use systems of a given tier, meant to be administered by regulating bodies such as national AI safety institutes or a standards body for AI rather than by the developers themselves. Developer safety frameworks [4], [5], [6] and regulators [7], [8] say what a developer must do as capability rises. MAIAM says who downstream may have that capability. The paper contributes:

1. a purpose-indexed tier structure (learning, exploration, frontier) that maps onto the actual constituencies who need AI access, rather than onto release artifacts or compute budgets alone (Section 3);
2. an explicit tier-assignment scheme along capability, autonomy, domain-consequence, and usage/safety axes, with a discussion of who adjudicates tier boundaries (Section 4);
3. an analysis of why safety evaluation of a released base model does not transfer to its open-weight derivatives, so that the correct object of safety reasoning is the downstream ecosystem (Section 5);
4. a two-dimensional safety framework combining the Accessibility Boundary defined here with the Capability Boundary proposed in the SAFE AI Operating Point (SAIOP) [9] (Section 7); and

5. a compact mathematical statement of proportionate accessibility and tier assignment (Section 8).

Tiered access does not eliminate misuse, and we do not claim it does. What it does is keep the safety levers that exist after release, which publishing frontier weights destroys, while leaving broad access in place where the risk is low. MAIAM makes AI offerings organized, manageable, and safe through clear classification of availability, accessibility, and accountability, and it limits the scope of dangers that cannot be contained under today's open-source, open-weights model. Once access is tiered, the traceability and accountability that Tiers 2 and 3 impose give AI safety problems a definite scope, which is what makes them manageable.

## 2 Related Work

### 2.1 Release gradients and structured access

Solaiman [1] characterizes generative-model release as a gradient from fully closed through gradual or staged release, hosted access, cloud or API access, downloadable weights, and fully open, and argues that each point trades off control against openness. Shevlane [2] introduces structured access: providing capabilities through controlled interfaces so that the developer can enforce use policies, monitor, and revoke. Seger et al. [3] evaluate open-sourcing highly capable foundation models and recommend against unrestricted release for the most capable systems, proposing staged release and gated access as alternatives that preserve many open-source objectives. All three are decisions a developer makes about its own interface. MAIAM differs in three ways: it tiers by user purpose and safety (learning, exploration, frontier work), it puts tier administration with regulating bodies rather than developers, and it pairs accessibility with an operating safe-limit ceiling as defined by SAIOP.

### 2.2 The open-weights debate

Eiras et al. [10] survey risks and opportunities across the development pipeline and argue for stage-appropriate governance. The U.S. NTIA report on dual-use foundation models with widely available weights [11] concluded in 2024 that evidence did not yet justify restrictions on open weights but recommended building the capacity to monitor and act as capabilities grow. Industry positions are not consistent: Anthropic and OpenAI favor closed frontier models, while NVIDIA, Meta [12], and Hugging Face favor open source and open weights.

### 2.3 Safety removal in open-weight models

A growing empirical literature shows that safety behavior installed by fine-tuning is shallow and cheaply reversible when weights are available. Qi et al. [13] show that fine-tuning on a few adversarial examples, or even on benign data, degrades safety alignment. Lermen et al. [14] undo safety training in Llama 2-Chat 70B with LoRA for under $200 of compute; Gade et al. [15] do the same for the 13B model; Yang et al. [16] demonstrate "shadow alignment" with roughly one hundred examples; and Zhan et al. [17] remove RLHF protections from a closed model through its fine-tuning API, showing that the problem is not exclusive to open weights but is far cheaper and irreversible when weights are public.

### 2.4 Capability thresholds and frontier regulation

Frontier developers have adopted capability-indexed safety frameworks: Anthropic's Responsible Scaling Policy [4], OpenAI's Preparedness Framework [5], and Google DeepMind's Frontier Safety Framework [6] each define capability thresholds that trigger stronger security and deployment controls. Regulators have used training-compute thresholds as a proxy: the EU AI Act presumes general-purpose models trained with more than $10^{25}$ floating-point operations to present systemic risk [7], and U.S. Executive Order 14110 set reporting thresholds at $10^{26}$ [8]. All of these say what developers must do as capability rises and little about who downstream may obtain that capability. The Accessibility Boundary fills that gap, and it can reuse the same thresholds and evaluations for tier assignment. SAIOP adds that a capability threshold by itself does not stop AI from running rogue; a safe operating limit is needed, set by five dimensions of AI safety: (1) Capability, (2) Behavior, (3) Society, (4) Governance, and (5) Ethics [9].

## 3 The Multi-Tiered AI Accessibility Model

**Principle 1 (Proportionate accessibility).**

The breadth and unconditionality of access to an AI system should be inversely related to its capability, autonomy, and potential for consequential harm, and conditioned on its demonstrated safety.

Two mistakes follow from ignoring this principle. One is to let every gain in capability arrive with unrestricted access attached. The other is to let worry about frontier systems justify locking down systems that pose little risk. Table 1 summarizes the three tiers and Fig. 2 shows the resulting capability–access gradient.

***Table 1:*** *The three tiers of the Multi-Tiered AI Accessibility Model.*

| | Tier 1: AI for Learning | Tier 2: AI for Exploration | Tier 3: Frontier AI |
|---|---|---|---|
| Purpose | Education, training, AI literacy | Research, development, experimentation | Highest-capability, highest-consequence work |
| Capability | Deliberately bounded | Substantial; below frontier thresholds | At or near the frontier |
| Access | Broad; anonymous use acceptable | Registration, identity, usage terms scaled to capability | Structured access for vetted organizations and purposes |
| Weights | Open weights appropriate | Open or gated depending on sub-tier | Not publicly downloadable |

| | Tier 1: AI for Learning | Tier 2: AI for Exploration | Tier 3: Frontier AI |
|---|---|---|---|
| Traceability | Not required | Identified users; usage recorded | Full: identity, logging, provenance |
| Post-release levers | Largely unnecessary | Monitoring and revocation where gated | Full: monitor, throttle, revoke, patch, retire |
| Accountability | Low stakes; minimal | Identified users and organizations | Contractual and institutional |

### 3.1 Tier 1: AI for Learning

Tier 1 systems exist to teach. Their users are students, educators, trainees, and members of the public who want to learn how AI systems work, how to use them, how to judge their output, how to build on them, how they affect society, and where they fail. None of this requires the most capable model available. A small or distilled model with transparent behavior delivers most of the pedagogical value at a fraction of the misuse potential, so Tier 1 systems should be as open as they can be: open weights, open training recipes where feasible [18], permissive licenses. Their capability is bounded by design rather than by a policy applied afterwards.

### 3.2 Tier 2: AI for Exploration

Tier 2 serves researchers, developers, businesses, and advanced learners who need capability beyond what Tier 1 provides: application development, scientific experimentation, engineering, business-process automation, safety and alignment research, specialized model evaluation, and agent experimentation. Capability in this tier is substantial but below the frontier thresholds defined in Section 4. Access controls scale with capability within the tier. Lower Tier 2 systems may be broadly available, possibly with open weights under licenses that impose use restrictions. Higher Tier 2 systems may require registration, identity verification, acceptance of use policies, and usage monitoring, and may be offered as gated downloads or hosted access rather than as fully open weights. These controls are not there to stop legitimate experimentation. They exist so that, as capability rises, someone identifiable is accountable for its use, that use is recorded, and there is a way to intervene if it is misused. Structured access for auditors and red-teamers [19], [20] is required in this tier and in Tier 3 alike; it is a cross-cutting access class rather than a tier of its own.

### 3.3 Tier 3: Full-Scale Frontier AI

Tier 3 comprises the most capable systems: those with advanced reasoning, coding, planning, or autonomous-agent capabilities whose misuse or malfunction could have significant real-world consequences. Application scenarios include defense, aviation, finance, and critical infrastructure. This is where the model parts company with open-release philosophy. That less capable systems are broadly accessible is no reason for a frontier system to be downloadable by anyone, modifiable without restriction, or redistributable at will. Access to Tier 3 capability should be provided according to legitimate need, organizational responsibility, technical safeguards, and the consequences of misuse, through hosted interfaces or structured access agreements, with identity, logging, and revocation. Every Tier 3 access is then traceable to an accountable party. Tier 3 has a ceiling as well as a floor: the Capability Boundary of Section 7, above which a system should not be built or deployed under any access regime.

## 4 Tier-Assignment Criteria

A tiered model is only as useful as its boundaries are operational. We propose that tier assignment be determined jointly along four axes, and that a system's tier is the highest tier indicated by any axis.

### 4.1 Capability axis

Capability can be estimated from three complementary indicators, in increasing order of fidelity and cost:

- *Training compute*, which is measurable before training and is already used as a regulatory proxy [7], [8], [21]. It is coarse and its correlation with capability drifts as algorithmic efficiency improves [22], so it should set a floor, not a ceiling.
- *Benchmark performance* on general and domain-specific evaluations, subject to the known problems of contamination and saturation.
- *Dangerous-capability evaluations* [23], [24] that probe uplift in biological, chemical, and cyber domains, persuasion, self-proliferation, and autonomous replication. These are the indicators that should dominate the Tier 2/Tier 3 boundary, consistent with the capability thresholds in existing developer frameworks [4], [5], [6].

### 4.2 Autonomy axis

Two systems of equal raw capability can pose very different risks depending on how autonomously they act. We propose a five-level ordinal scale, informed by but distinct from the autonomy levels of Morris et al. [25]: (0) the system only responds to direct prompts; (1) it uses tools within a single turn under human confirmation; (2) it executes multi-step plans with human checkpoints; (3) it operates for extended periods with delegated authority over external actions; (4) it can acquire resources, spawn sub-agents, or modify its own deployment. Autonomy levels 3–4 push a system toward Tier 3 even when its capability metrics alone would indicate Tier 2, because autonomy multiplies the consequence of any single misuse or malfunction [26], [27].

### 4.3 Domain-consequence axis

Systems specialized for or deployed in high-consequence domains (biological design, critical infrastructure control, defense, aviation, financial market operations, mass communication at scale) warrant a higher tier than general-purpose systems of equivalent capability. This axis captures the fact that the same capability has different worst-case consequences in different domains.

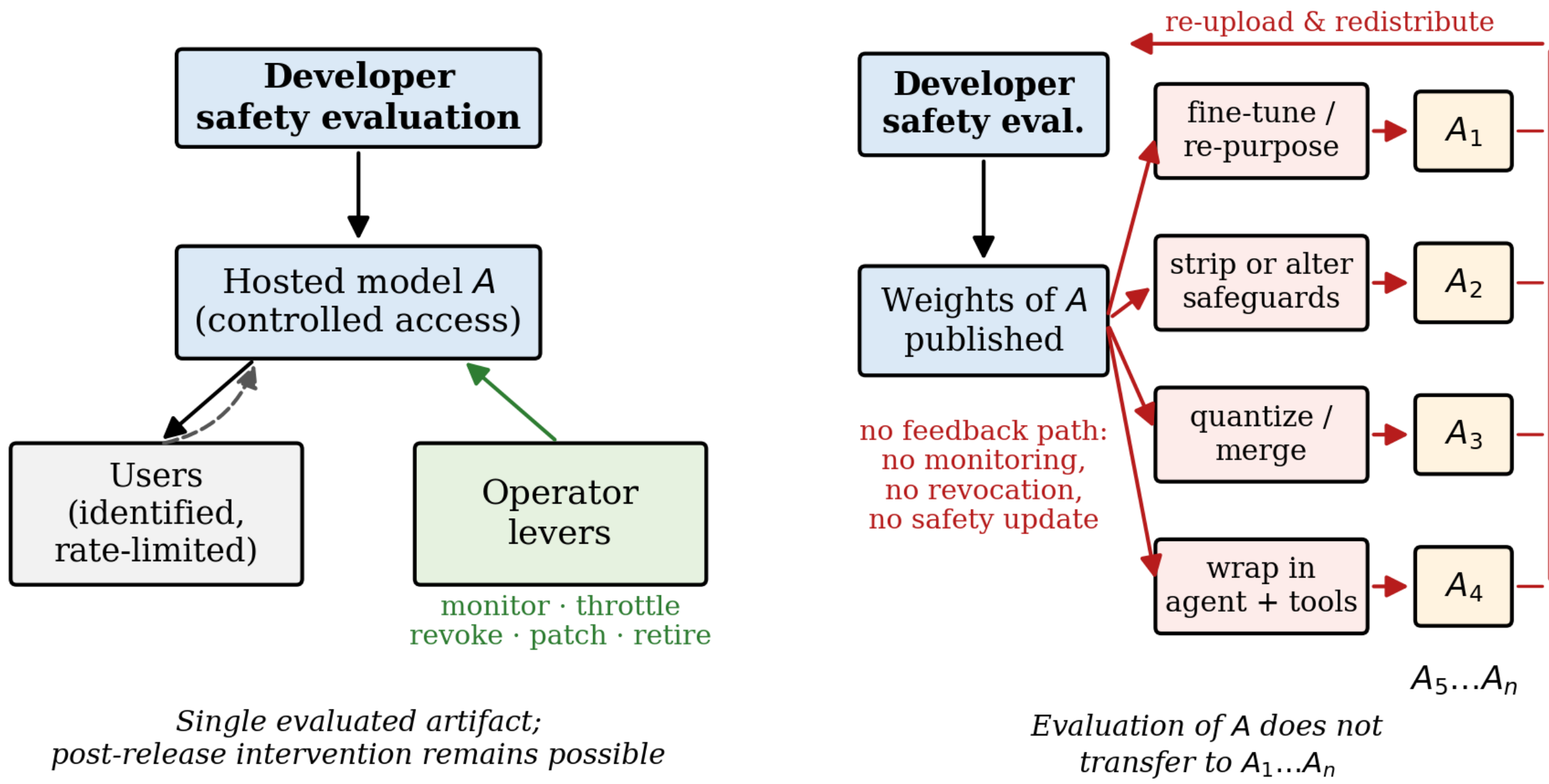

(a) Controlled or structured access

(b) Unrestricted open-weight release

***Figure 1:*** *Two post-release regimes. (a) Under controlled or structured access, a single evaluated artifact A is served to identified users, and the operator retains levers to monitor, throttle, revoke, patch, and retire. (b) Under unrestricted open-weight release, the evaluated artifact A is transformed into derivatives* $A_1, \ldots, A_n$ *by fine-tuning, safeguard removal, quantization and merging, and agentic wrapping; derivatives are re-uploaded and further derived. No feedback path returns to the developer, and the evaluation of A does not transfer to its derivatives.*

### 4.4 Usage and safety axis

The three axes above describe what a system can do. The tiers are named for purpose, so assignment must also take in what a system is for and how safely it has been shown to behave. We fold these into a usage/safety indicator with two parts.

*Declared usage.* The purpose for which a system is offered places a floor on its tier: a system offered for general education and literacy is a Tier 1 candidate; a system offered for development, research, or experimentation is at least Tier 2; a system offered for operational deployment in the domains of Section 4.3, or for autonomous operation, is at least Tier 3. Declared usage is checked against the other three axes: a system declared for education whose capability or autonomy indicates Tier 2 or 3 is assigned the higher tier, because purpose cannot lower the tier that capability demands.

*Safety posture.* Within a tier, the conditions of access depend on the system's demonstrated safety: whether it has passed the dangerous-capability evaluations relevant to its tier, whether its safeguards survive adversarial fine-tuning (the test that matters for any open-weight release, Section 5), whether provenance and usage records are maintained, and whether it has an incident history. A system that meets its tier's safety requirements receives the broadest access the tier permits; a system that has not yet demonstrated them, or whose safeguards have been shown removable, receives the narrowest, or is provisionally assigned the next tier up until the evidence is supplied.

This is the axis a regulating body actually works with. It assigns a tier from the developer's declaration of purpose and its evidence of safety, and it audits both.

### 4.5 Who assigns tiers

Three arrangements are available and are not mutually exclusive. First, developers may self-assign under published frameworks, as they do today [4], [5], [6]; this is fast but lacks independent verification. Second, regulators may set statutory thresholds, as the EU has done for systemic-risk models [7]; this is authoritative but slow and coarse. Third, an independent evaluation body, of the kind envisaged by national AI safety institutes, may certify tier assignments against published evaluations and the usage/safety declarations of Section 4.4. We recommend the third arrangement for defining tier levels and for providing governance over the accessibility of tiered AI systems. MAIAM is written for such bodies: national AI safety institutes within their jurisdictions and, across jurisdictions, a standards body for AI accessibility doing for access tiers what the IETF does for Internet protocols, that is, publishing the tier criteria, the evidence each tier requires, and the access conditions attached to it.

## 5 The Control Problem of Unrestricted Open Weights

This section gives the reason Tier 3 systems should not be released as open weights. Fig. 1 contrasts the two regimes.

### 5.1 Benefits of open weights

Open weights deliver real benefits that any restriction must be weighed against [3], [10], [28]: they enable local and private deployment, customization and fine-tuning for underserved languages and domains, reproducible research, independent auditing, reduced dependence on a small number of providers, lower barriers to entry, and a distribution of capability and economic value beyond incumbent firms. Tiers 1 and 2 exist to keep these benefits. What follows concerns the frontier only.

### 5.2 Loss of post-release levers

A developer who hosts a model can monitor usage, impose rate limits, detect anomalous activity, revoke access, update safeguards, investigate abuse, and discontinue the model [2]. Once weights are downloaded, each of these levers is lost with respect to that copy. Another party can fine-tune the model for a different purpose, combine it with other models, alter or remove safety mechanisms, quantize it for cheaper deployment, embed it in an autonomous agent with tool access, and redistribute the result. None of this needs the developer's cooperation or even its knowledge. So the question stops being whether the released model is safe and becomes whether the downstream ecosystem can still be controlled at all.

### 5.3 Evaluation does not transfer to derivatives

Pre-release safety evaluation is performed on a specific artifact. The empirical literature reviewed in Section 2 shows that this evaluation is fragile under modification: safety fine-tuning can be removed with a few hundred examples and a few hundred dollars of compute [13], [14], [15], [16]. Consequently a released model $A$ that passed evaluation may spawn derivatives $A_1, \dots, A_n$ that would not have passed, and there is no mechanism by which the derivatives are evaluated before they are used. Under closed access, an equivalent attack through a fine-tuning API [17] is detectable and revocable by the provider; under open weights it is neither. What was evaluated and what is deployed drift apart, and the drift widens with each generation of derivatives.

### 5.4 Proliferation and provenance

Once weights are public, the number of derivative systems is bounded only by the size of the community that can fine-tune them. Public model repositories host millions of models, a substantial share of which are fine-tunes, merges, and quantizations of a comparatively small number of base releases. That is a measure of how much innovation open release enables. It is also a measure of how fast one frontier release becomes a family of systems whose provenance, safeguards, tool integrations, and deployments its creator knows nothing about. Redistribution makes it worse: a derivative can be created, uploaded, downloaded, modified, and re-uploaded again and again, and attribution and accountability thin out at every step.

### 5.5 Why this matters more at the frontier

For low-capability systems this trade-off is acceptable. The worst an unsafeguarded derivative can do is limited, and the benefits of openness win. As capability, autonomy, and domain consequence rise, so does the expected cost of having no post-release lever left. The tiered model does not settle the trade-off once and for all. It settles it differently at different points on the capability axis, which is all Principle 1 asks.

## 6 Comparative Analysis

Table 2 sets the two regimes side by side on the dimensions that matter after release. The contrast is between capability-invariant access, meaning the same access regime whatever the capability, and capability-indexed access, rather than between "open" and "closed."

***Table 2:*** *Unrestricted open-weight release versus multi-tiered accessibility.*

| Dimension | Unrestricted open weights | Multi-tiered accessibility |
|---|---|---|
| Educational AI | Broadly accessible | Broadly accessible (Tier 1) |
| Research AI | Broadly accessible | Accessible according to capability and purpose (Tier 2) |
| Frontier AI | Downloadable | Structured access (Tier 3) |
| Modification and fine-tuning | Unrestricted | Depends on tier |
| Redistribution | Unrestricted | Constrained at higher tiers |
| User identification | Typically absent | Introduced as capability rises |
| Traceability | None after download | Recorded at Tiers 2–3 |
| Safety updates | Infeasible after proliferation | Feasible while systems remain controlled |
| Monitoring | Lost once weights proliferate | Retained for controlled systems |
| Accountability | Diffuse across derivatives | Attached to identifiable parties at higher tiers |
| Proliferation | Rapid, unbounded | Deliberately constrained at higher tiers |
| Basis of control | Release mode | Capability, autonomy, domain, usage/safety |
| System-wide safety | Degrades with capability | Designed to hold as capability rises |

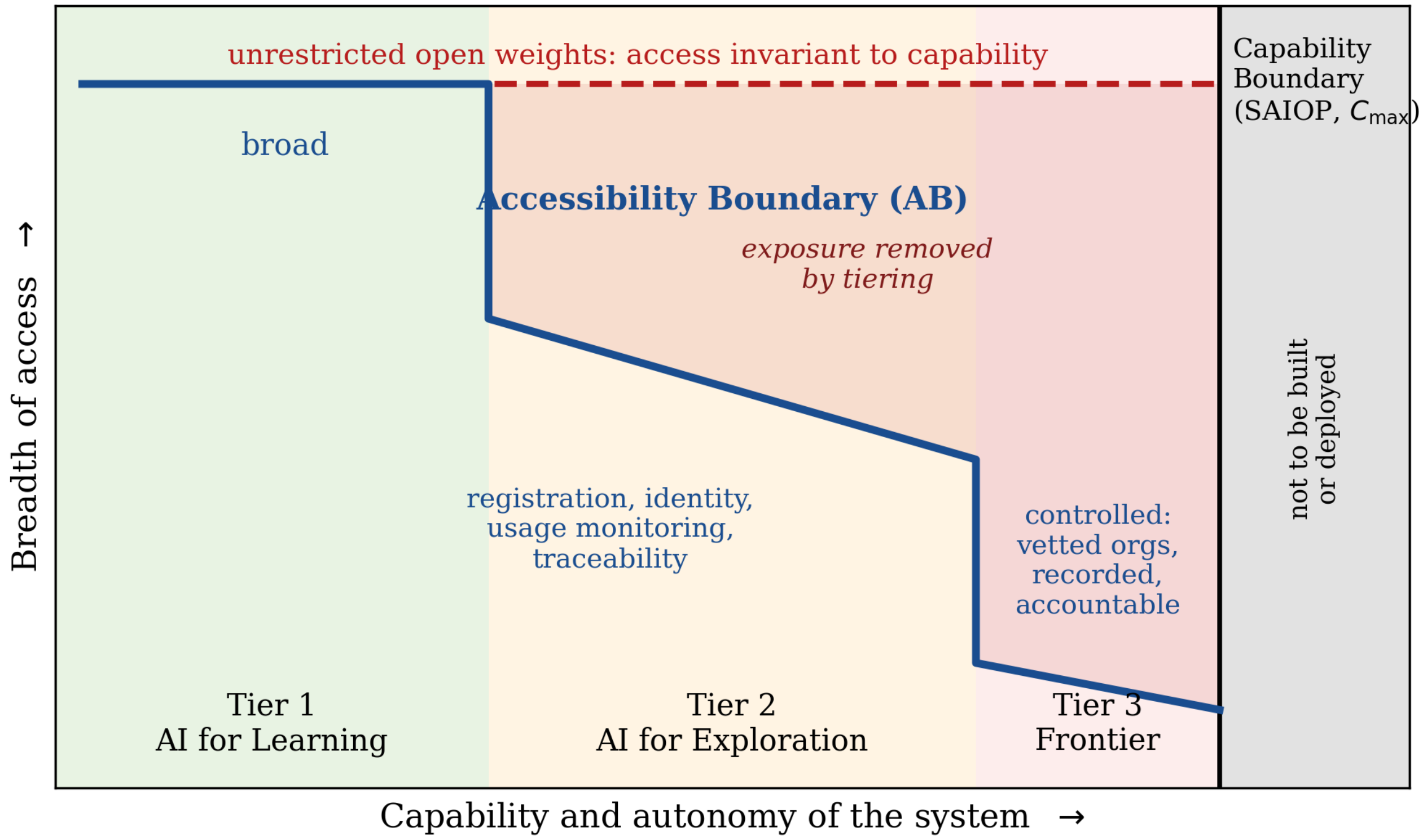

***Figure 2:*** *The two-dimensional safety framework (schematic; axes are ordinal and the curves are illustrative, not measured). The Capability Boundary (vertical line) bounds what should be built. Within it, the Accessibility Boundary (solid) prescribes decreasing breadth of access as capability and autonomy rise across the three tiers. Unrestricted open-weight release (dashed) holds access constant regardless of capability; the shaded region is the exposure that tiering removes.*

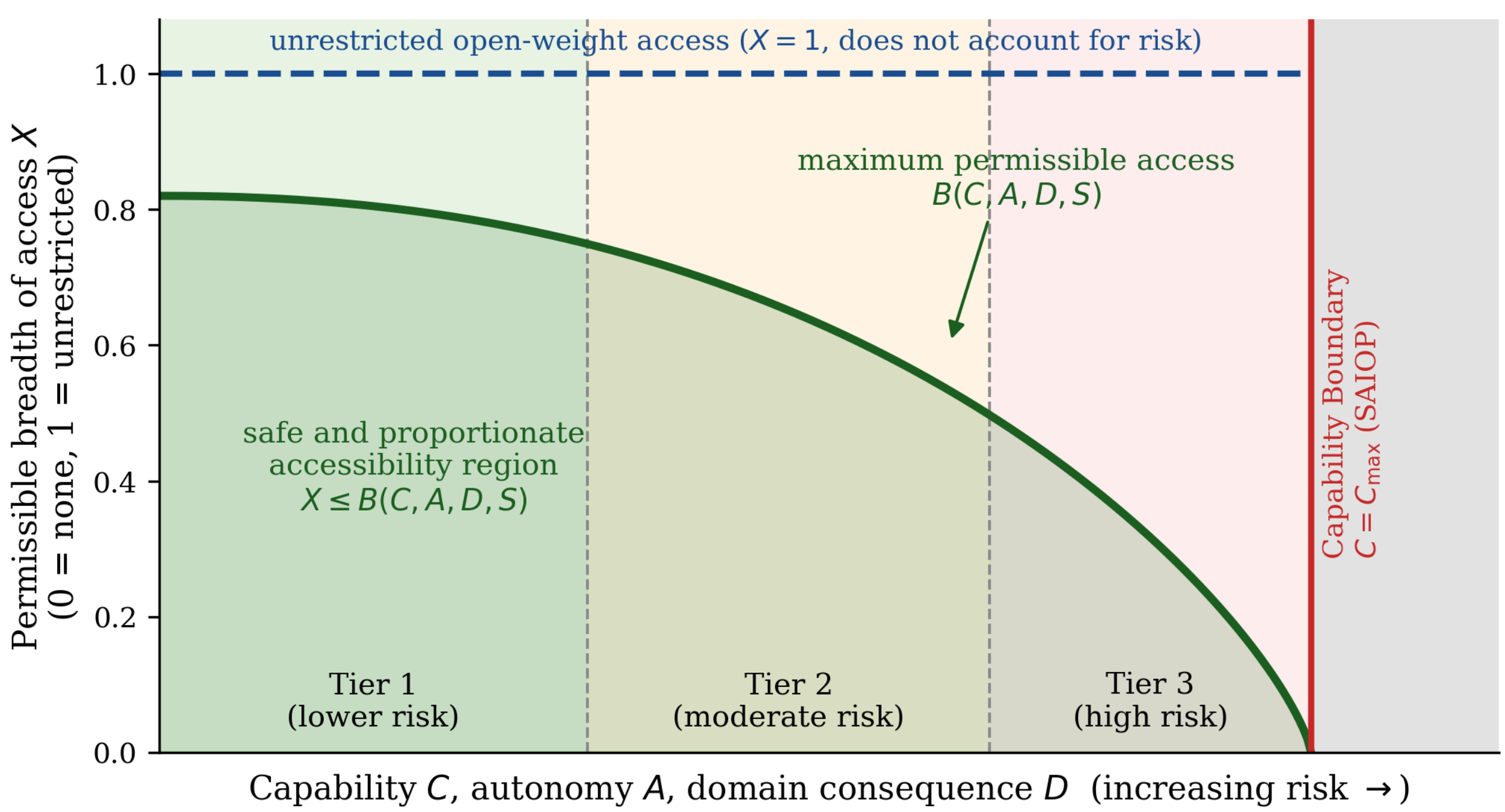

***Figure 3:*** *Risk-constrained accessibility as a function of capability. The permissible breadth of access $X$ is bounded above by $B(C, A, D, S)$, which decreases monotonically with capability $C$, autonomy $A$, and domain consequence $D$, and reaches zero at the Capability Boundary $C_{max}$. Unrestricted open-weight access ($X = 1$, dashed) does not account for risk. The shaded region beneath the curve is the safe and proportionate accessibility region of Section 8.*

Seven things follow from adopting the tiered regime.

1. *Educational access is preserved.* Students, educators, and the public are not excluded from AI; Tier 1 remains fully open.
2. *Innovation is preserved.* Researchers and developers obtain capability matched to their needs in Tier 2, and vetted organizations obtain frontier capability in Tier 3 through structured access.
3. *Uncontrolled proliferation is reduced at the frontier.* If frontier weights are not public, the population of uncontrolled copies and derivatives is bounded.
4. *Safety updates remain possible.* A vulnerability or newly discovered dangerous capability in a controlled system can be patched or the system retired; in a population of independent copies it cannot.
5. *Traceability and accountability become practical.* Higher-tier access is associated with identifiable users and institutions and a record of use.
6. *Failures are compartmentalized.* A safety failure in a Tier 1 system does not confer Tier 3 capability; the tiers create separation between capability classes.
7. *Controls scale with capability.* The regime does not impose frontier-grade restrictions on educational models, nor does it allow frontier capability to inherit the access regime of educational models.

## 7 Relationship to the SAFE AI Operating Point

The SAFE AI Operating Point (SAIOP) [9] proposes an upper bound on AI capability and autonomy beyond which humans cannot retain effective control, and argues that systems should be operated below that point. SAIOP answers one question: how capable and autonomous a system may be allowed to become. MAIAM answers the other: who may obtain a given level of capability, and on what terms. Together they define two boundaries (Fig. 2):

**Definition (Capability Boundary).**

The Capability Boundary (CB) is the maximum capability and autonomy that should be built or deployed, as determined by SAIOP. It governs the safety limits not to be crossed.

**Definition (Accessibility Boundary).**

The Accessibility Boundary (AB) is the function that maps each capability level below the CB to the breadth and conditions of access appropriate to it, as determined by the tier-assignment criteria of Section 4.

Under open-weight release the AB is flat. Access does not change with capability, and the CB is the only safety instrument left. Under the tiered model the AB slopes down, and the region between the two curves, which is frontier-class capability in unaccountable hands, disappears. Neither boundary is enough on its own. A CB without an AB leaves every capability below the ceiling available to everyone, including capabilities that are dangerous in some hands and harmless in most. An AB without a CB allows controlled access to systems that nobody should be operating. Developer safety frameworks [4], [5], [6] already combine the two for a single company, since a capability threshold triggers both deployment limits and a commitment to pause if safeguards fall behind. We separate the two and apply them across the ecosystem, under regulating bodies, so that accessibility is governed on its own account and not as a by-product of release mode. Fig. 3 shows the resulting curve.

## 8 Summary of Mathematical Framework

The model can be stated in a few lines. Let a system be characterized by capability $C$, autonomy $A$, domain consequence $D$, and a usage/safety indicator $S$ (Section 4.4). Let $X \in [0,1]$ denote the breadth of access granted to the system, with $X = 0$ meaning no access and $X = 1$ meaning unrestricted, anonymous downloadability.

**Capability Boundary.** Let $C_{\max}$ be the maximum permissible capability determined by SAIOP. The Capability Boundary is

$$C \leq C_{\max}.$$

Systems with $C > C_{\max}$ should not be built or deployed, irrespective of access regime.

**Tier assignment.** Each axis induces a tier indicator $T_C(C)$, $T_A(A)$, $T_D(D)$, $T_S(S) \in \{1,2,3\}$ according to the criteria of Section 4. The tier of a system is the highest tier indicated by any axis:

$$T = \max\{\, T_C(C),\ T_A(A),\ T_D(D),\ T_S(S) \,\}.$$

**Accessibility Boundary.** Let $B(C, A, D, S) \in [0,1]$ denote the maximum permissible breadth of access. The Accessibility Boundary is

$$X \leq B(C, A, D, S).$$

Principle 1 requires $B$ to be non-increasing in each risk variable and non-decreasing in demonstrated safety:

$$\frac{\partial B}{\partial C} \leq 0, \qquad \frac{\partial B}{\partial A} \leq 0, \qquad \frac{\partial B}{\partial D} \leq 0, \qquad \frac{\partial B}{\partial S} \geq 0,$$

with $B(C_{\max}, \cdot, \cdot, \cdot) = 0$. This is the model's prediction in one line: as capability, autonomy, or consequence goes up, permissible access goes down (Fig. 3).

**Safe deployment region.** The two-dimensional safety structure is the set

$$\mathcal{S} = \{\, (C, A, D, S, X) \colon C \leq C_{\max},\ X \leq B(C, A, D, S) \,\}.$$

**Risk-implicit form.** Equivalently, if $R(C, A, D, K)$ denotes expected downstream risk as a function of capability, autonomy, domain consequence, and the effective control strength $K$ that the access regime provides (identity, logging, revocation, patching), and $R_{\max}$ is the risk tolerance set by the regulating body, permissible accessibility may be defined implicitly by

$$B(C, A, D, S) = \max\{\, X \colon\ R(C, A, D, K(X, S)) \leq R_{\max} \,\},$$

where $K$ decreases as $X$ increases (broader access means weaker control; at $X = 1$ every lever of Section 5 is lost) and increases with $S$. Capability and accessibility are therefore separate variables. The first says what a system can do; the second says on what terms that capability can be had or used.

## 9 Conclusion

Who may have an AI system is a different question from how capable that system may be, and it deserves its own governance. MAIAM answers it with three purpose-indexed tiers, assigns systems to them along capability, autonomy, domain-consequence, and usage/safety axes, and keeps access broad where risk is low while making it progressively more traceable and accountable as capability and risk rise. With the SAFE AI Operating Point it forms a two-part framework: an upper safety-limit Capability Boundary on what can be built, and an Accessibility Boundary on who may obtain it. The framework is meant for regulating bodies, national AI safety institutes and a standards body for AI accessibility, which would assign tiers from declared usage and demonstrated safety and administer the access conditions that go with each.

The empirical premise underneath all of this is that pre-release evaluation does not survive open-weight release. Derivatives the evaluation never saw can be produced cheaply, and nothing remains to correct them. That is why we believe a multi-tiered accessibility model far exceeds the safety of an open-source, open-weight scenario, and why AI safety can be managed, constrained, and accounted for through MAIAM.

**Limitations.** MAIAM is a governance framework rather than a technical mechanism. It depends on institutions to assign tiers and run access, and on an evaluation science that is still young. Restricting frontier access concentrates capability in fewer hands; a broad Tier 2 and a structured rather than closed Tier 3 for vetted researchers and auditors manage that cost without removing it. Weights leak, and some developers in some jurisdictions will publish regardless, so the framework raises the price and lowers the frequency of frontier proliferation rather than ending it, and it needs the regulating bodies that run it to coordinate across borders. None of this changes the basic point. When access does not vary with capability, the only instrument left is the capability ceiling.

## Author Contributions

C. K. Toh proposed the Multi-Tiered AI Accessibility Model and is the first author. V. R. Jain developed the tier-assignment criteria, the control-problem analysis, the comparative and discussion sections, the mathematical framework, and the figures. Both authors wrote and revised the manuscript. V. R. Jain is the corresponding author.